\documentclass[10pt,twocolumn,letterpaper]{article}

\usepackage[pagenumbers]{cvpr} % To force page numbers, \eg for an arXiv version

\usepackage{graphicx}
\usepackage{amsmath}
\usepackage{amssymb}
\usepackage{booktabs}

\usepackage{multirow}
\usepackage{caption}
\usepackage{subcaption}
\usepackage{comment}
\usepackage{makecell} % make multirows in fixed length cell
\usepackage{floatrow} % align caption vertically
\usepackage{placeins}

\usepackage[utf8]{inputenc}
\usepackage{multirow}
\usepackage{amsfonts}
\newcommand{\Lagr}{\mathcal{L}}
\usepackage{mathrsfs}

\usepackage[pagebackref,breaklinks,colorlinks]{hyperref}

\usepackage[capitalize]{cleveref}
\crefname{section}{Sec.}{Secs.}
\Crefname{section}{Section}{Sections}
\Crefname{table}{Table}{Tables}
\crefname{table}{Tab.}{Tabs.}

\def\cvprPaperID{67} % *** Enter the CVPR Paper ID here
\def\confName{CVPR}
\def\confYear{2023}

\begin{document}

\title{Robustness to distribution shifts of compressed networks for edge devices}

\author{Lulan Shen, Ali Edalati, Brett Meyer, Warren Gross, James J. Clark\\
McGill University\\
Montreal, Quebec, Canada\\
{\tt\small lulan.shen@mail.mcgill.ca, james.j.clark@mcgill.ca}
%\and
}

\maketitle

%%%%%%%%% ABSTRACT
\begin{abstract}
It is necessary to develop efficient DNNs deployed on edge devices with limited computation resources. However, the compressed networks often execute new tasks in the target domain, which is different from the source domain where the original network is trained. It is important to investigate the robustness of compressed networks in two types of data distribution shifts: domain shifts and adversarial perturbations. In this study, we discover that compressed models are less robust to distribution shifts than their original networks. Interestingly, larger networks are more vulnerable to losing robustness than smaller ones, even when they are compressed to a similar size as the smaller networks. Furthermore, compact networks obtained by knowledge distillation are much more robust to distribution shifts than pruned networks. Finally, post-training quantization is a reliable method for achieving significant robustness to distribution shifts, and it outperforms both pruned and distilled models in terms of robustness.
\end{abstract}

%%%%%%%%% BODY TEXT
\section{Introduction}
\label{sec:intro}

Over the past few years, the field of deep learning has witnessed a remarkable surge in the development of large-scale models \cite{simonyan2015deep, chen2017deeplab}. The deeper and wider architectures of neural networks improve their performance but demand considerable storage and computational resources, preventing them from being deployed on edge devices \cite{chen2021addernet, he2019filter}, \eg, mobile phones, smartwatches, IoT devices, \etc. This is because these devices often have limited resources in terms of memory, processing power, and energy consumption, which makes it challenging to run large models efficiently. 

To overcome this challenge, various network compression methods have been developed to reduce the memory footprint and computational complexity while preserving the model's accuracy. These techniques can be categorized into different main approaches, including network pruning (\ie, weight pruning \cite{han2015learning}, filter pruning \cite{Molchanov2017pruning, Yu2017Accelerating}), knowledge distillation (KD) \cite{hinton2015distilling}, quantization \cite{rastegari2016xnornet, qin2020bnn, hubara2016binarized, courbariaux2016binaryconnect}, low-rank approximation \cite{iandola2016squeezenet, han2016deep, Denton2014exploiting} and compact network design \cite{Chollet2017Xception, howard2017mobilenets}. 

%Deep neural networks (DNNs) \cite{LeNet5} have achieved superior performance in many computer vision tasks such as image classification \cite{krizhevsky2012imagenet, simonyan2015deep, zeiler2013visualizing}, object detection \cite{ren2016faster}, semantic segmentation \cite{long2015fully, chen2017deeplab}, and human face verification \cite{wen_discriminative_2016}, as well as in applications such as speech recognition \cite{graves_framewise_2005} and natural language processing \cite{collobert2011natural}. The deeper and wider architectures of neural networks improve their performance but demand considerable storage and computational resources, preventing them from being deployed on edge devices \cite{chen2021addernet, he2019filter}, \eg, mobile phones, smartwatches, \etc. Additionally, high-end Graphics Processing Unit (GPU) cards are typically required to perform the computations involved in neural networks, which leads to high power consumption and additional hardware requirements \cite{chen2021addernet}. Therefore, it is essential to design efficient neural networks which can run with limited computation resources on edge devices. 

%DNNs are typically over-parameterized, leading to a waste of memory and computation resources \cite{denil2014predicting}. 

However, it is challenging to make compressed neural networks perform well in the target domain. Typically, the original networks are trained on a source domain, \eg, ImageNet dataset \cite{ImageNet}, but compressed networks execute new tasks in a target domain, which has the same classes as the source domain but lacks labeled training data. Therefore, there is always a distribution shift between the source and target domains due to various factors, \eg, pose positions, image quality, sensor noise, domain-specific characteristics, \etc. \cite{Wang2018DAsurvey}. In this paper, we specifically define data distribution shifts as situations where data samples used for training and testing are sourced from different datasets or domains. 

Iofinova \etal \cite{Iofinova2022how} investigated the transfer performance of pruned convolutional neural networks (CNNs) by evaluating the image classification accuracy of these networks in transfer learning tasks. Specifically, they fine-tuned pre-trained and compressed models on ImageNet on different downstream tasks, which contained classes different from those in ImageNet. However, the study did not evaluate the performance of pruned CNNs or other types of compressed models under distribution shifts between training and application distributions with the same classes.

Further research is required to investigate the performance of compressed models under distribution shifts, particularly when target domain data is scarce and unlabeled. In such cases, it is challenging to train a standalone network, which emphasizes the need to address distribution shifts to ensure the model's performance in the target domain. Furthermore, deploying neural networks on edge devices is prevalent in real-life scenarios, and the effect of distribution shifts on compressed neural networks cannot be ignored. Therefore, it is crucial to study the effect of different compression techniques on neural network robustness to distribution shifts, enabling people to prioritize a compression technique that prioritizes model robustness. 

We empirically investigate the robustness of compressed object classification networks to input data distribution shift, using three commonly-used network compression techniques: filter pruning, KD, and post-training quantization. Our objective is to identify if any of them have an advantage over the others in terms of handling distribution shifts. This study examines two distribution shifts: domain shifts due to input images acquired under varying environmental conditions, and adversarial attacks, where input data is perturbed to alter the network's classification results. 

Our findings reveal that the performance of compressed models in target domains degrades as their compression ratio increases. Additionally, compressing from a larger base model instead of a smaller one to achieve a particular size of the compact model results in a less robust performance in distribution shifts. Importantly, we conclude that post-training quantization is the most effective compression technique for handling distribution shifts, especially in the case of domain shifts. Particularly, when compressing the model to one-quarter of its original size is sufficient, post-training quantization proves to be a favorable approach.

To summarize, our contributions are:
\begin{itemize}
\vspace{-7pt}
    \item Evaluating adversarial and domain shift robustness of compressed ResNets using pruning, KD, and quantization on the Office-31 dataset.
    \vspace{-7pt}
    \item Investigating the effect of compression rate on the robustness of pruned and distilled models.
    \vspace{-7pt}
    \item Evaluating the robustness of compressed models under a wide range of attacks in different strengths. 
\end{itemize}

%-------------------------------------------------------------------------

\section{Related Works}
\subsection{Model Compression}
%\subsubsection{Network Pruning}
\textit{\textbf{Network Pruning:}} The comprehensive literature reviews on compressing DNNs are presented in \cite{mishra2020survey, cheng2020survey}. Weight pruning is a simple compression method that removes redundant weights of neural networks \cite{meng2020pruning}, which converts a dense network into a sparse
one \cite{han2015learning}. Filter pruning is a structured way to reduce computation costs in CNNs without sparse networks or special hardware \cite{li2017pruning} which is usually required for compressing and accelerating convolutional layers \cite{han2016EIE}. In addition, filter pruning is faster and more efficient than weight pruning by removing entire channels instead of pruning a single neuron connection.

%\subsubsection{Knowledge Distillation}
\textit{\textbf{Knowledge Distillation:}} Knowledge distillation \cite{hinton2015distilling, suau2019filter, Gou_2021} is a popular compression method where knowledge is transferred from a large ``teacher" model to a smaller ``student" model. Mimicking the teacher's performance, the lightweight student model achieves competitive results in tasks like visual recognition and natural language processing. The knowledge type, distillation algorithm, and teacher-student architectures are crucial in this process. The vanilla KD \cite{hinton2015distilling} uses the teacher model's logits to train the student model with a distillation loss, which aligns the predictions between the two models.

%\subsubsection{Quantization}
\textit{\textbf{Quantization:}} Neural network quantization is a technique that can reduce the model size and lower computation overheads by decreasing the precision of parameter representation without sacrificing accuracy \cite{Gholami2022Quantization}, \ie, converting a 32-bit float precision to an 8-bit integer representation. Han \etal \cite{han2016deep} introduced an influential technique that combines pruning, quantization, and Huffman coding to compress DNNs. Network binarization is another popular quantization approach, which constrains the weights and activations of neural networks to be binary values \cite{rastegari2016xnornet,qin2020bnn,hubara2016binarized}.

\subsection{Adversarial Attacks}

\begin{figure*}[h!]
  \begin{subfigure}{0.24\textwidth}
    \centering
    \includegraphics[width=0.75\linewidth]{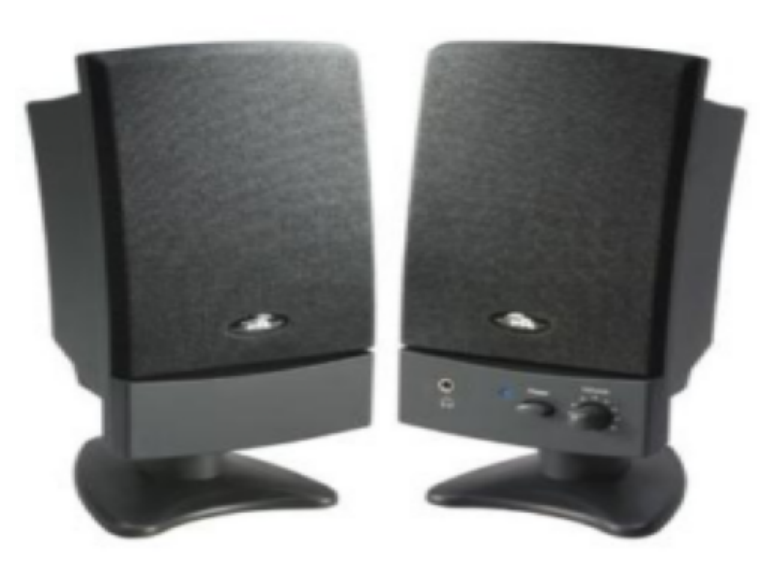}
    \caption{Original: \texttt{Speaker}}
  \end{subfigure}
  \begin{subfigure}{0.24\textwidth}
    \centering
    \includegraphics[width=0.75\linewidth]{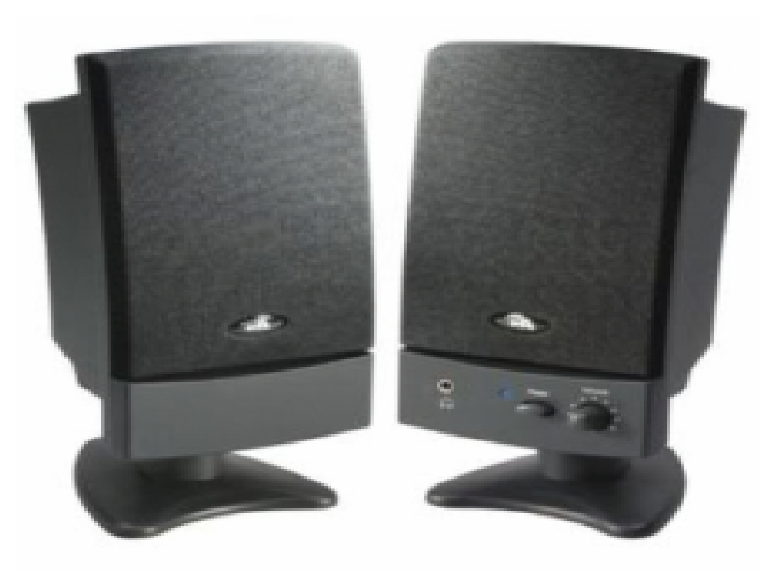}
    \caption{C\&W: \texttt{computer}}
  \end{subfigure}
  \begin{subfigure}{0.24\textwidth}
    \centering
    \includegraphics[width=0.75\linewidth]{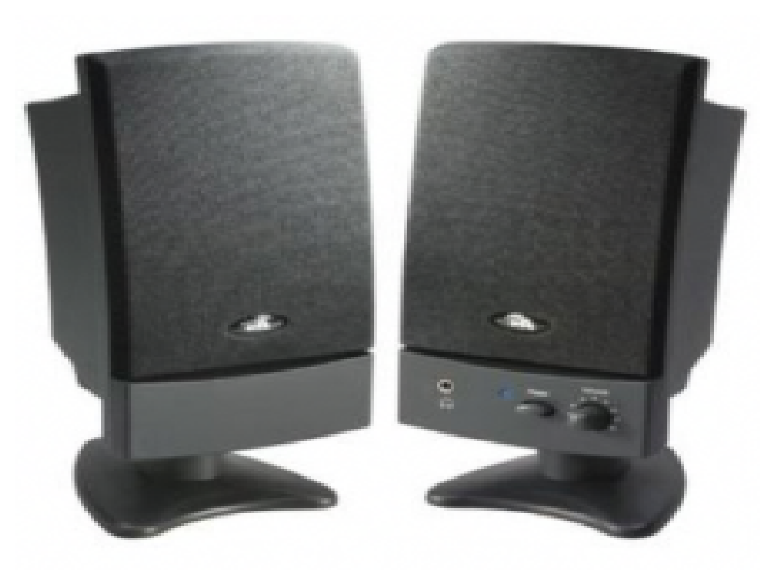}
    \caption{DeepFool: \texttt{chair}}
  \end{subfigure}
  \begin{subfigure}{0.24\textwidth}
    \centering
    \includegraphics[width=0.75\linewidth]{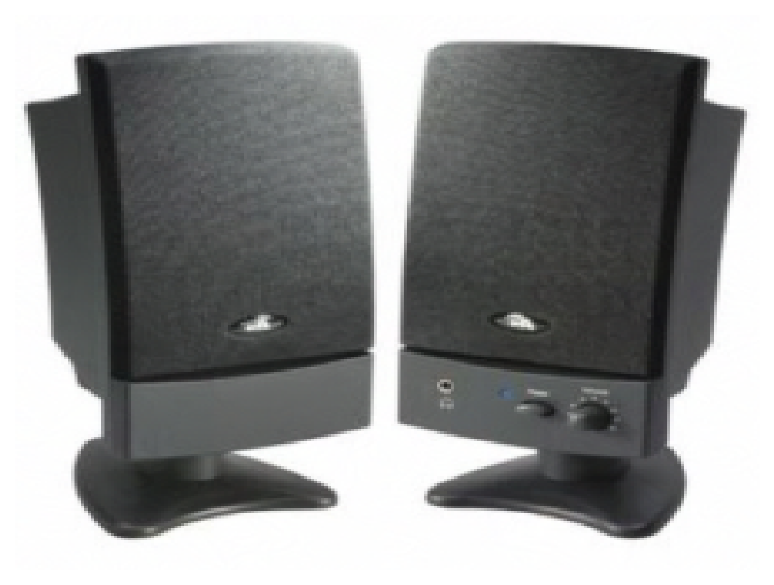}
    \caption{PGD: \texttt{lamp}}
  \end{subfigure}
  \begin{subfigure}{0.24\textwidth}
    \centering
    \includegraphics[width=0.75\linewidth]{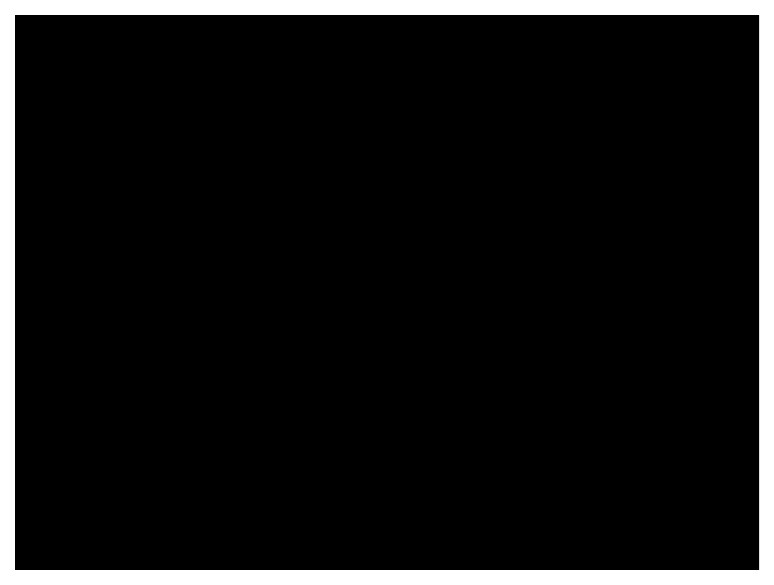}
    \caption{No perturbation}
  \end{subfigure}
  \begin{subfigure}{0.24\textwidth}
    \centering
    \includegraphics[width=0.75\linewidth]{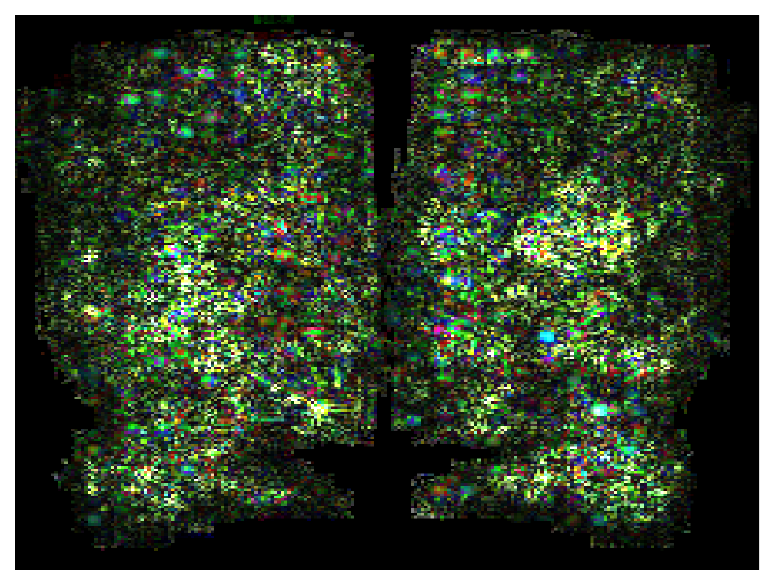}
    \caption{C\&W perturbation}
  \end{subfigure}
  \begin{subfigure}{0.24\textwidth}
    \centering
    \includegraphics[width=0.75\linewidth]{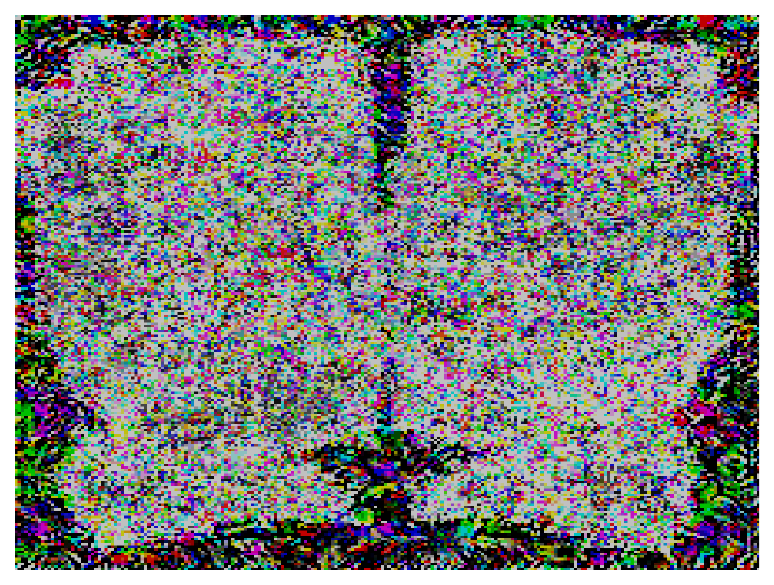}
    \caption{DeepFool perturbation}
  \end{subfigure}
  \begin{subfigure}{0.24\textwidth}
    \centering
    \includegraphics[width=0.75\linewidth]{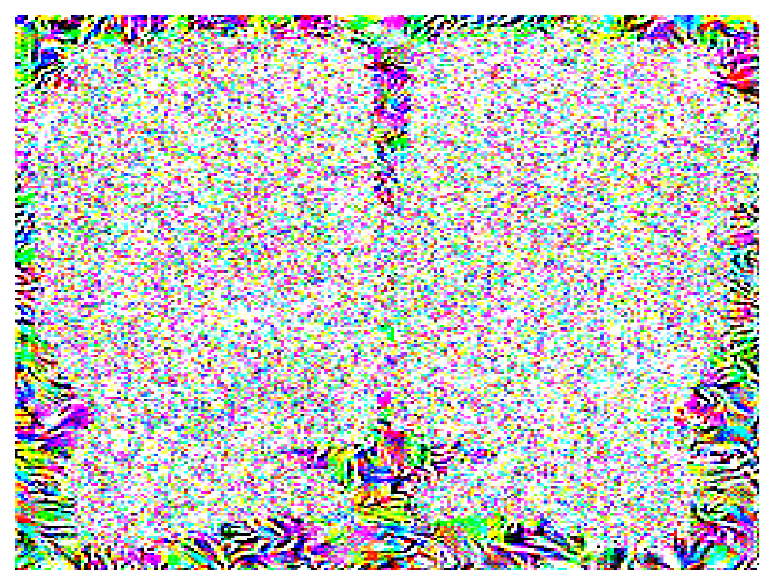}
    \caption{PGD perturbation}
  \end{subfigure}
  \caption{Subplots (a)-(d) show misclassified images of a speaker from the Amazon domain by ResNet-50 under C\&W, DeepFool, and PGD adversarial attacks. Subplots (e)-(h) show the corresponding perturbations generated under attacks, magnified by a factor of 500.}
  \label{fig:adv}
\end{figure*}

In recent years, there has been an increasing amount of research aimed at developing techniques to deceive neural networks. These techniques, known as adversarial attacks, involve making malicious yet subtle changes in the input to fool the network (See \cref{fig:adv}) \cite{goodfellow2014explaining,madry2018towards,rony2019decoupling,carlini2017adversarial,moosavi2016deepfool}. Alternately, various techniques have been developed to enhance the robustness of neural networks against such attacks \cite{madry2018towards,qin2019adversarial,bashivan2021adversarial,zhang2019towards}. Also, several works have introduced techniques to improve the adversarial robustness of compressed models by combining adversarial training \cite{madry2018towards} and compression \cite{Goldblum2020Adversarially,ye2019adversarial,Gui2019Model,Wijayanto2019towards}. 

Moreover, empirical and theoretical analyses presented in \cite{guo2018sparse} suggested that the sparsity and non-linearity of DNNs improve their adversarial robustness. Later, Jordao \etal \cite{Jordao2021on} empirically showed that pruning improves adversarial robustness without investigating the effect of the pruning rate. Shumailov \etal  \cite{shumailov2019compress} studied the transferability of adversarially generated data samples between compressed and uncompressed models in addition to the adversarial robustness of the compressed models. However, they only investigated pruning, quantization, and a few gradient-based adversarial attacks. 

By measuring the adversarial robustness of binarized neural networks, Galloway \etal \cite{galloway2018attacking} revealed that quantization causes \textit{gradient masking} \cite{papernot2017practical}, which improves the robustness of quantized models against gradient-based attacks. In addition, Lin \etal \cite{lin2018defensive} concluded that the 
robustness of quantized models is improved against small perturbations but reduced against large perturbations. They also introduced \textit{defensive quantization} to improve the adversarial robustness of quantized models.

% To the best of our knowledge, our work is one of the most comprehensive studies on the domain shift and adversarial robustness of compressed neural networks through pruning, quantization, and knowledge distillation.

% \cite{Wijayanto2019towards} briefly investigated the effect of quantization on the adversarial robustness of the compressed AlexNet \cite{krizhevsky2012imagenet}, which also proposed an adversarial-aware compression framework based on combining pruning, quantization, and DEFLATE encoding.

% For instance, \cite{Ye2019Adversarial,Vemparala2021Adversarial} proposed frameworks that  combine pruning and adversarial training \cite{madry2017towards} to develop compressed yet robust networks. \cite{Gui2019Model} introduced a unified framework for adversarial training of compressed networks by pruning, quantization, and factorization. 
\section{Methodology}
\label{sec:meth}

\subsection{Model Compression}
%We compressed different versions of ResNets \cite{he2016deep} into smaller versions using three major compression techniques to reduce the size of the models.% while maintaining their performances. %The selected technique is explained briefly in the following subsection. 
We use three primary compression techniques to reduce the size of various ResNets \cite{he2016deep}, discussed below.

\subsubsection{Pruning}
We apply $\Lagr1$-norm Filter Pruning ($\Lagr1$-FP) \cite{li2017pruning} for pruning. This technique assumes filters or neurons with smaller weights are less important in the model. Each filter, denoted as $\mathcal{F}_{i\in [1,I]}$, has $J$ weight elements in a model with $I$ filters. The importance score for each filter is calculated using:
\begin{equation}
\vspace{-5pt}
Score(\mathcal{F}_{i})= ||\mathcal{F}_{i}|| = \sum_{j=1}^{J} |\mathcal{F}_{i,j}|.
\end{equation}
$\Lagr1$-FP measures the sum of absolute weights for all filters in a well-trained model and sorts the filters based on the relative importance score of filters in each layer. The least important filters in each layer are then pruned to achieve a desired compression ratio. Finally, the pruned model is trained again on the source domain.

\subsubsection{Knowledge Distillation}
We use vanilla KD \cite{hinton2015distilling}, where we train the large teacher model on the source domain and compute the logits for each class using the softmax output layer, as shown below:
\begin{equation}
    p_i (\tau) = \frac{e^{t_i/\tau}}{\sum_{j=1}^N e^{t_j/\tau}}, \quad q_i (\tau) = \frac{e^{s_i/\tau}}{\sum_{j=1}^N e^{s_j/\tau}},
\end{equation}
where $t_i$ and $s_i$ are the logits generated by the classification layers of the teacher and student network, respectively, given a data sample for the class $i$; $N$ is the number of classes; and $\tau$ is the temperature parameter, which is set to 1 by default. The higher $\tau$ value softens the probability distribution output over classes. The ``soft targets" of the teacher and student model are outputs of the softmax layers: $\mathbf{p}=(p_1,..,p_N)$ and $\mathbf{q}=(q_1,..,q_N)$ respectively. We use the class probabilities generated by the teacher model as ``soft targets" for student model training, transferring the generalization ability of the larger teacher model to the compact student model.

The objective function is calculated as below:
\begin{equation}
\vspace{-5pt}
\Lagr_{KD}=CrossEntropy(\mathbf{p}, \mathbf{q})=-\sum_{i=1}^{N}{p_i^\tau \log q_i^\tau},
\end{equation}
\begin{equation}
% \vspace{-5pt}
\Lagr_{CE}=CrossEntropy(\mathbf{y}, \mathbf{q})=-\sum_{i=1}^{N}{y_i \log q_i},
\end{equation}
\begin{equation}
\Lagr_{total} = \alpha \tau^2 \Lagr_{KD} + (1-\alpha)\Lagr_{CE},
\end{equation}
where $\mathbf{y}$ is the one-hot vector of the ground-truth label, and $\alpha$ is the weighting hyper-parameter. $\Lagr_{CE}$ measures the classification loss of the student model, while $\Lagr_{KD}$ encourages the student model's predicted probabilities to match the teacher model's probabilities.

\subsubsection{Quantization}
We also apply network quantization, which reduces the precision of computations and weight storage by using lower bit-widths instead of floating-point precision. We choose post-training static quantization (PTSQ) \cite{Jacob2018Quantization}, which is one of the most common and fastest quantization techniques in practice. This technique determines the scales and zero points prior to inference. Specifically, we quantize the 32-bit weights (\eg, $w \in [\alpha, \beta]$) and activations of the trained baseline models to 8-bit integer values (\eg, $w_q \in [\alpha_q, \beta_q]$). The quantization process is defined as
\begin{equation}
    w_q = \text{round} \left(\frac{1}{s} w + z \right),
\end{equation}
where $s$ is the scale, and $d$ is the zero-point, defined as
\begin{equation}
    s = \frac{\beta-\alpha}{\beta_q - \alpha_q}, \quad z = \text{round} \left(\frac{\beta \alpha_q - \alpha \beta_q}{\beta - \alpha} \right).
\end{equation}

\subsection{Domain Shifts}
\begin{figure}[ht]
    \centering
    \includegraphics[width=0.5\textwidth]{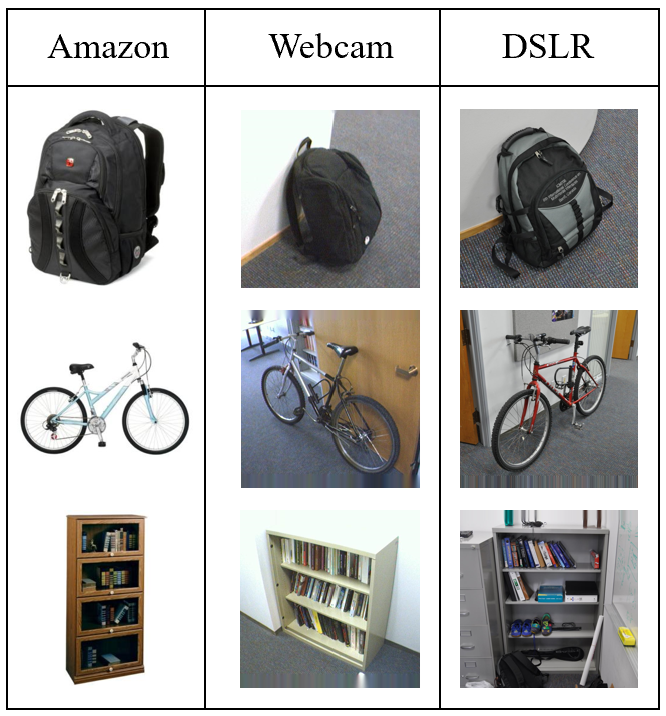}
    \caption{Sample images of backpack, bike, and bookcase from Amazon, Webcam, and DSLR domain of the Office-31 \cite{Saenko2010AdaptingVC}.}
    \label{fig:img_office31}
\end{figure}

In this study, we evaluate our approach on the Office-31 dataset \cite{Saenko2010AdaptingVC} (shown in \cref{fig:img_office31}). This dataset is a widely used benchmark for domain adaptation research. It consists of 4,110 images of 31 object classes from an office environment, divided into three image domains: \textit{Amazon} ($\mathcal{D}_A$), \textit{Webcam} ($\mathcal{D}_W$), and \textit{DSLR} ($\mathcal{D}_D$). The $\mathcal{D}_A$ domain contains 2,871 medium-resolution images with studio lighting conditions, downloaded from the Amazon website; the $\mathcal{D}_W$ domain has 795 low-resolution images with significant noise and white balance artifacts, captured with a simple webcam; the $\mathcal{D}_D$ domain contains 498 low-noise high-resolution images captured with a digital SLR camera in realistic environments with natural lighting conditions \cite{Saenko2010AdaptingVC}. Each domain is split into 90\% train data and 10\% validation data using a random seed of 1, and the model is trained on the train data. We report the validation accuracy on the source domain and the testing accuracy on the entire target domain. We perform six domain shifts in total, where we train on one Office-31 domain and test on one of the other two domains.

\subsection{Adversarial Attacks}
\label{sec:adv}
We evaluate the robustness of models against seven commonly used adversarial attacks, briefly discussed below.
\begin{itemize}
%vspace{-7pt}
    \item \textbf{Fast Gradient Sign Method (FGSM)}: 
    This method generates an adversarial sample, $x_{adv}$, by adding the scaled gradient of the loss function ($\nabla_x J(x,y,\theta)$) to the input image ($x$) \cite{goodfellow2014explaining}. \Cref{eq:fgsm} shows how this method generates the adversarial data, where $y$ represents the image label, $\theta$ represents the model parameters, and $\epsilon$ is the perturbation strength.
    \begin{equation}
        \label{eq:fgsm}
        x_{adv} = x + \epsilon \text{sign}(\nabla_x J(x,y,\theta))
    \end{equation}
    \item \textbf{Projected Gradient Decent (PGD)}: This method \cite{madry2018towards} is an iterative version of FGSM. The input image $x_t$ is updated to $x_{t+1}$ using 
    \begin{equation}
        \label{eq:pgd}
        x_{t+1} = \prod_{x+S} (x_t + \delta \text{sign}(\nabla_x J(x,y,\theta))),
    \end{equation}
    where $\delta$ is the step size, and $S$ is the set of 
    possible perturbations. 
    % The attack is considered successful if an adversarial image is generated within the maximum number of iterations. 
    \item \textbf{DeepFool}: This method approximates the model with a linear classifier and measures the minimum distance required to reach the decision boundary of each class. Finally, the adversarial sample is obtained by adding the minimum distance to the input image \cite{moosavi2016deepfool}. 
    \item \textbf{Decoupled Direction and Norm (DDN)}: This attack is an iterative approach that refines the noise added to the input image in each iteration to make it adversarial.
    At iteration $i$, the adversarial input image, $x_i$, is generated as $x_i=x+\eta_i$, where $\eta_i$ is the noise with a norm of $\sigma_i$. If $x_i$ is adversarial, the norm of the next iteration noise is decreased ($\sigma_{i+1}=\sigma_i(1-\epsilon)$). Otherwise, the norm of the next noise is increased ($\sigma_{i+1}=\sigma_i(1+\epsilon)$). This process repeats until the minimum required perturbation is found \cite{rony2019decoupling}.
    \item \textbf{Carlini and Wagner (C\&W)}: 
    This is considered one of the strongest adversarial attacks as it aims to find the minimum perturbation needed to change the predicted class of an input image. C\&W reformulates the mentioned optimization problem to find the adversarial sample by using an optimizer \cite{carlini2017adversarial} like Adam \cite{Kingma2015Adam}. 
    
    \item \textbf{Elastic-Net Attacks to
    DNNs (EAD)}: This technique uses Iterative Shrinkage-Thresholding Algorithm \cite{Beck2009AFI} to solve the same optimization problem that C\&W is trying to solve. 
    \item \textbf{Salt\&Pepper}: A non-gradient-based attack that repeatedly adds Salt \& Pepper noise to the input to fool the model.
\end{itemize}

%The strength of each of these attacks can be controlled with a hyper-parameter, known as the perturbation strength ($\epsilon$). Several perturbation strengths are investigated for each adversarial attack to find the corresponding $\epsilon$ for a light and heavy attack according to the accuracy drop of ResNet-18. 
We evaluate the strength of adversarial attacks by varying the perturbation strength ($\epsilon$) and assessing both light and heavy attacks based on the accuracy drop of ResNet-18. All compressed models obtained through pruning and KD are evaluated against all discussed attacks under both light and heavy attacks, including gradient-based and non-gradient-based ones. However, the PTSQ framework does not support backward path calculations, so we only evaluate the quantized models against EAD, C\&W, and Salt\&Pepper.

%----------------------------------------------------------
\section{Experimental Setup} \label{sec:experiments}
We initialize ResNets \cite{he2015deep} with various depths (\ie, ResNet-18, -34, -50, -101, and -152) using pre-trained parameters on ImageNet \cite{ImageNet} from PyTorch \cite{paszke_pytorch_2019}. 
To obtain our baseline models (\ie, \textit{baseline-A}, \textit{baseline-W}, and \textit{baseline-D}), we fine-tuned these models on each source domain (\ie, {$\mathcal{D}_A$, $\mathcal{D}_W$, $\mathcal{D}_D$}). The input images are resized to $224 \times 224$ and the same data transforms as the pre-trained models are used. We use a stochastic gradient descent (SGD) optimizer \cite{SGD} with a momentum of 0.9, a fixed learning rate (LR) of 1e-4, a batch size of 8 for \{$\mathcal{D}_A, \mathcal{D}_W$\} and of 16 for $\mathcal{D}_D$, and trained for 100 epochs. We select the fine-tuned models with the best validation accuracy as our baseline models. These baseline models are then evaluated directly on the target domain. %even though the best validation accuracy on the source domain may not guarantee the best test accuracy on the target domain. In real-life applications, the target domain is usually unknown, and only the model which obtains the best validation or training accuracy is saved for later use. Baseline models are then evaluated directly on the target domain. 

To obtain compact models, we compress the baseline models using $\Lagr1$-FP pruning \cite{li2017pruning}, KD \cite{hinton2015distilling}, and PTSQ \cite{Jacob2018Quantization}. First, we apply $\Lagr1$-FP pruning to the baseline models, using a single sensitivity hyper-parameter to control the pruning ratios in each ResNet block. The compressed models are then re-trained from scratch on the source domain for 100 epochs. We use an SGD optimizer with a momentum of 0.9, a weight decay of 1e-4, and a batch size of 8. For ResNet-50 and ResNet-152, we used a fixed LR of 1e-4, while for ResNet-18, ResNet-34, and ResNet-101, we used an adaptive LR with a base LR of 1e-3, divided by 10 at epoch 70. For KD, we use the larger baseline models (\ie, already fine-tuned on the source domain) as our teacher models and the pre-trained smaller ResNet models (provided by PyTorch) as our student models. We set $\alpha$ as 0.2 and $\tau$ as 10. We train the student models for 100 epochs using an SGD optimizer with a momentum of 0.9, a weight decay of 1e-3, and a batch size of 8. We use a step LR scheduler, which divided the base LR by 5 at epoch 70. For PTSQ, we use the built-in quantization modules provided by PyTorch, which allows us to fuse modules, calibrate the model using the training data to determine the appropriate scale factor, and finally quantize the weights and activation in the model. We choose hyper-parameters for the pruned and distilled models that maximize their validation accuracy.
%(summarized in \cref{table:l1fp_para} in the Appendix) 

%For PTSQ, we fuse activations into preceding layers
%https://pytorch.org/blog/quantization-in-practice/#post-training-static-quantization-ptq
%

To evaluate the performance of compressed models and baselines against adversarial attacks, we use the fine-tuned checkpoints on Amazon (\ie, ResNets baseline-A and its compressed models). Then we use Foolox\footnote{\href{https://github.com/bethgelab/foolbox}{https://github.com/bethgelab/foolbox}} \cite{rauber2017foolbox,rauber2017foolboxnative} to generate an adversarial test dataset for each model, using the original test dataset obtained by splitting 0.1 of randomly selected data samples from $\mathcal{D}_A$. The batch size for the adversarial experiments is set to 8. We use default hyper-parameters for most of the attacks, except for EAD, C\&W, and Salt\&Pepper, which require 100, 100, and 400 steps, respectively. Also, we set the step size of C\&W to 0.001. %Note that the input images undergo the ImageNet transformation with a lower and upper bound of 0 and 1, respectively.

We implement our models and experiments using PyTorch and run them on V100LGPU. Hyper-parameters are chosen using a basic grid search that tunes batch size, base LR, step size, and gamma for the step LR scheduler, weight decay of the SGD optimizer, $\alpha$, and $\tau$ for KD. %The selected hyper-parameters for baselines and compressed models are summarized in \cref{table:other_hyperparameter} in the Appendix.

%----------------------------------------------------------
\section{Results and discussion}
\subsection{Domain Shifts} \label{sec:domain_shifts}
\begin{table*}[hbt!]
\footnotesize
\caption{The \textit{validation} accuracies (\%) of ResNets models on each \textit{source} domain of the Office-31 dataset. The baseline (uncompressed) ResNets are obtained after fine-tuning the pre-trained model on the ImageNet dataset. The pruned models are obtained using the $\Lagr1$-FP method with different pruning ratios, and the distilled/student models are obtained using teacher networks of different sizes.}
\centering
\begin{tabular}{ c|c|c|c|c|c|c|c} \toprule
    Base/Teacher Model & Compression Method & \# Params (M) & Compression Rate (\%) & Amazon & Webcam & DSLR & Avg Acc\\ \midrule
    \multirow{2}{*}{ResNet-18} & - & 11.19 & - & 89.32 & 89.87 & 93.88 & 91.02\\ 
    & FP & 4.51 & 59.70 & 89.68 & 87.34 & 87.76 & 88.26\\ \midrule
    \multirow{4}{*}{ResNet-34} & - & 21.30 & - & 90.75 & 89.87 & 93.88 & 91.50\\
    & FP & 11.19 & 47.46 & 85.41 & 89.87 & 93.88 & 89.72\\ 
    & FP & 4.51 & 78.83 & 75.44 & 78.48 & 79.59 & 77.84\\ 
    & KD & ResNet18 (11.19) & 47.46 & 90.04 & 91.14 & 97.96 & 93.05\\\midrule
    \multirow{6}{*}{ResNet-50} & - & 23.57 & - & 90.04 & 89.87 & 97.96 & 92.62\\ 
    & FP & 21.30 & 9.63 & 92.17 & 89.87 & 97.96 & 93.33\\
    & FP & 11.19 & 52.52 & 87.90 & 86.08 & 93.88 & 89.29\\ 
    & FP & 4.51 & 80.87 & 75.44 & 83.54 & 81.63 & 80.20\\  
    & KD & ResNet34 (21.30) & 9.63 & 90.04 & 89.87 & 97.96 & 92.62\\
    & KD & ResNet18 (11.19) & 52.52 & 90.39 & 91.14 & 97.96 & 93.16\\ \midrule
    \multirow{4}{*}{ResNet-101} & - & 42.56 & - & 91.81 & 91.14 & 97.96 & 93.64\\
    & FP & 23.57 & 44.62 & 90.04 & 89.87 & 97.96 & 92.62\\ 
    & FP & 11.19 & 73.71 & 86.12 & 88.61 & 97.96 & 90.90\\ 
    & FP & 4.51 & 89.40 & 72.95 & 84.81 & 77.55 & 78.44\\  
    \midrule
    \multirow{6}{*}{ResNet-152} & - & 58.21 & - & 90.75 & 87.34 & 95.92 & 91.34\\
    & FP & 42.56 & 26.89 & 91.46 & 91.14 & 95.92 & 92.84\\ 
    & FP & 23.57 & 59.51 & 89.68 & 88.61 & 97.96 & 92.08\\ 
    & FP & 11.19 & 80.78 & 82.56 & 88.61 & 95.92 & 89.03\\ 
    & FP & 4.51 & 92.25 & 68.68 & 81.01 & 85.71 & 78.47\\ 
    & KD & ResNet18 (11.19) & 80.78 & 90.04 & 91.14 & 95.92 & 92.37\\ \bottomrule
\end{tabular}
\label{table:resnet_office_source}
\end{table*}

\begin{table*}[htb!]
\footnotesize
\caption{The \textit{test} accuracies (\%) of ResNets on \textit{target} domains of the Office-31 dataset other than what they were trained on.} %``A", ``W", and ``D" represent the domain of Amazon, Webcam, and DSLR.}
\centering
\begin{tabular}{c|c|c|c|c|c|c|c|c|c}
\toprule
 Base/Teacher Model & Compression Method & \# Params (M) & A $\rightarrow$ W & A $\rightarrow$ D & W $\rightarrow$ A & W $\rightarrow$ D & D $\rightarrow$ A & D $\rightarrow$ W & Avg Acc\\\midrule
\multirow{2}{*}{ResNet-18} & - & 11.19 & 64.15 & 60.84 & 50.05 & 97.39 & 47.43 & 92.70 & 68.76 \\
& FP & 4.51 & 57.36 & 59.64 & 47.39 & 96.99 & 44.73 & 91.32 & 66.24\\
\midrule
\multirow{4}{*}{ResNet-34} & - & 21.30 & 67.80 & 68.27 & 54.67 & 98.59 & 52.01 & 92.70 & 72.34\\ 
& FP & 11.19 & 48.93 & 51.61 & 46.22 & 95.78 & 43.66 & 89.69 & 62.65\\ 
& FP & 4.51 & 16.73 & 14.06 & 16.68 & 63.86 & 10.47 & 57.36 & 29.86\\ 
& KD & 11.19 & 62.52 & 63.25 & 48.49 & 99.00 & 52.79 & 95.47 & \textbf{70.25}\\ \midrule
\multirow{6}{*}{ResNet-50} & - & 23.57 & 68.93 & 71.89 & 64.25 & 99.00 & 60.95 & 94.84 & 76.64 \\
& FP & 21.30 & 69.43 & 72.69 & 61.52 & 98.39 & 59.53 & 94.34 & 75.98\\
& FP & 11.19 & 53.46 & 61.45 & 41.36 & 93.57 & 35.64 & 85.28 & 61.79 \\
& FP & 4.51 & 19.37 & 26.10 & 11.89 & 50.00 & 9.23 & 34.72 & 25.22\\
& KD & 21.30 & 57.74 & 59.04 & 56.59 & 98.80 & 58.18 & 97.11 & \textbf{71.24}\\
& KD & 11.19 & 57.99 & 57.63 & 52.18 & 99.40 & 48.78 & 95.47 & \textbf{68.58} \\\midrule
\multirow{4}{*}{ResNet-101} & - & 42.56 & 74.59 & 75.70 & 64.32 & 99.00 & 24.35 & 90.19 & 71.36\\ 
& FP & 23.57 & 62.89 & 63.45 & 54.10 & 98.59 & 55.80 & 91.19 & 71.00\\
& FP & 11.19 & 43.02 & 40.16 & 28.19 & 90.36 & 19.63 & 84.65 & 51.00\\
& FP & 4.51 & 19.37 & 17.27 & 12.11 & 57.83 & 5.18 & 32.58 & 24.10\\\midrule
\multirow{6}{*}{ResNet-152} & - & 58.21 & 74.97 & 74.70 & 63.97 & 98.39 & 63.76 & 95.72 & 78.59\\
& FP & 42.56 & 71.19 & 72.69 & 61.95 & 99.00 & 61.59 & 93.46 & 76.65\\
& FP & 23.57 & 65.41 & 66.47 & 55.13 & 99.00 & 43.02 & 86.79 & 69.30\\
& FP & 11.19 & 36.23 & 39.36 & 25.49 & 81.93 & 10.69 & 57.11 & 41.80\\
& FP & 4.51 & 17.86 & 15.26 & 9.83 & 50.00 & 4.40 & 23.02 & 20.06\\
& KD & 11.19 & 58.49 & 62.25 & 48.49 & 98.39 & 50.80 & 95.85 & \textbf{69.05}
\\\bottomrule 
\end{tabular}
\label{table:resnet_office_target}
\end{table*}

\begin{table*}[htb!]
    \centering
    \footnotesize
    \caption{Accuracies of ResNet baselines and quantized 8-bit ResNets trained on the Amazon domain and accuracies of quantized ResNets for light and heavy adversarial perturbations. Note that a ``-" in the compression method column indicates the \textit{baseline} model is used.}
    %\resizebox{\textwidth}{!}{%
    \begin{tabular}{c|c|c|ccc|cccccc}
    \toprule
    \multirow{2}{*}{\begin{tabular}[c]{@{}c@{}} Base \\ Model \end{tabular}} &
    \multirow{2}{*}{\begin{tabular}[c]{@{}c@{}} Compression \\ Method \end{tabular}} & 
    \multirow{2}{*}{\begin{tabular}[c]{@{}c@{}} Memory Size \\ (MB) \end{tabular}} & 
    \multirow{2}{*}{\begin{tabular}[c]{@{}c@{}} Source \\ (A) \end{tabular}} & 
    \multirow{2}{*}{\begin{tabular}[c]{@{}c@{}} Target \\ (W) \end{tabular}} & 
    \multirow{2}{*}{\begin{tabular}[c]{@{}c@{}} Target \\ (D) \end{tabular}} & 
    \multicolumn{2}{c}{EAD} & 
    \multicolumn{2}{c}{C\&W$_{L_2}$} & 
    \multicolumn{2}{c}{Salt\&Pepper}\\
    & & & & &  & $\epsilon=1$ & $\epsilon=10$ & $\epsilon=0.4$ & $\epsilon=4$ & $\epsilon=1$ & $\epsilon=15$ \\
    \midrule
     \multirow{2}{*}{ResNet-18} & - & 42.79 & 88.61 & 63.77 & 66.87 & 83.99 & 50.89 & 14.23 & 0.00 & 85.77 & 31.67 \\
     & PTSQ & 10.81 & 88.26 & 63.02 & 65.26 & 87.90 & 87.90 & 87.90 & 87.90 & 85.77& 44.84 \\ 
     \midrule
     \multirow{2}{*}{ResNet-34} & - & 81.42 & 90.39 & 64.28 & 68.27 & 81.49 & 49.47 & 14.95 & 0.71 & 84.70 & 50.89 \\ 
     & PTSQ & 20.55 & 89.32 & 64.28 & 67.47 & 89.32 & 88.97 & 88.97 & 88.97 & 87.54 & 46.98 \\
     \midrule
     \multirow{2}{*}{ResNet-50} & - & 90.26 & 90.04 & 72.45 & 75.30 & 84.34 & 48.04 & 13.17 & 0.00 & 87.19 & 42.35 \\ 
     & PTSQ & 23.26 & 87.19 & 71.45 & 72.69 & 87.90 & 88.26 & 87.90 & 87.54 & 86.83 & 58.01 \\
     \midrule
     \multirow{2}{*}{ResNet-101} & - & 163.03 & 90.39 & 75.09 & 76.71 & 83.99 & 57.30 & 12.10 & 00.36 & 87.90 & 64.41 \\ 
     & PTSQ & 42.17 & 89.68 & 74.21 & 74.50 & 87.19 & 87.54 & 88.26 & 86.48 & 82.92 & 61.92 \\
     \midrule
     \multirow{2}{*}{ResNet-152} & - & 223.01 & 90.75 & 74.84 & 76.91 & 87.54 & 61.92 & 15.30 & 00.36 & 89.68 & 65.12 \\ 
     & PTSQ & 57.81 & 88.26 & 73.46 & 75.90 & 88.97 & 88.97 & 89.32 & 88.61 & 86.12 & 58.72 \\
     \bottomrule
    \end{tabular}%
    %}
    \label{tab:adversarial_qt}
\end{table*}

The ResNet baselines are achieved after fine-tuning ResNets pre-trained models on the corresponding source domain. Then, we compress the baseline models using the $\Lagr1$-FP with different pruning rates and distill the baselines to different sizes of student models using vanilla KD. \Cref{table:resnet_office_source} records the validation accuracies of various baselines and compressed ResNets to the training distribution, which is one of the three domains in the Office-31 dataset. The results show that the baseline models perform well in the source domain they were trained on. However, as we compress the model with the filter pruning technique, the validation accuracy of the compressed model in the source domain gets worse as the pruning ratio increases. In contrast, the distilled models (\ie, the ResNet-18 student model) generally have better training performance than the model with the same structure and size (\ie, the ResNet-18 baseline model). This is expected since the larger teacher model transfers its generalization ability to the student model.

Next, we evaluate the baseline and compressed/distilled models on the target domains and report their accuracies in \cref{table:resnet_office_target}. The results show that both baseline and compressed models of ResNets perform poorly on the target domains and their performance decreases as the compression rate increases, indicating a lack of robustness to domain shifts. However, after pruning their ResNet baseline models to a particular size, the smaller baseline model has a higher test accuracy on the target domains than the larger baseline networks, which are less affected by domain shifts. For example, the compressed ResNet with 4.51M parameters pruned from ResNet-18 has an average accuracy of 66.24\% on target domains, while the pruned models with the same size obtained from the base models ResNet-34, ResNet-50, ResNet-101, and ResNet-152 have average accuracies of 29.86\%, 25.22\%, 24.10\%, and 20.06\% respectively. Similar patterns are observed for other compressed network sizes. This suggests that it is more beneficial to prune a smaller base model rather than a larger one to a particular size since its corresponding compressed model has better generalization performance on unseen domains.

However, the situation differs for networks compressed through KD. In this case, the student models obtained using KD can possess similar or even better performance in target domains compared to the small baseline network of the same size. For instance, consider the networks with a parameter count of 11.19M. The average accuracy for the unpruned baseline is 68.76\%. However, as shown in \cref{table:resnet_office_target}, the accuracy of student models with 11.19M is generally superior to that of the small baseline network, and the decline in performance when using larger teachers is minimal. 

Furthermore, \cref{tab:adversarial_qt} demonstrates the performance of quantized 8-bit ResNets trained on the Amazon domain. The quantization process reduces the size of the model to one-quarter of its original size without any significant loss or degradation of accuracy in target domains. On average, quantized ResNet models, compressed to 1/4 of their original size, experience only a slight decrease in their target-domain accuracy of about 1.2\%. In contrast, models pruned and distilled from baseline-A, even at a relatively modest pruning ratio of 50\% (\ie, compressing from ResNet-50 to a model with 11.19M parameters), experience a much larger decrease in target-domain accuracy of approximately 13\%. This suggests that quantized models exhibit significantly greater robustness to domain shifts compared to pruned and distilled models. However, further research should investigate the robustness of quantized 4-bit/2-bit integer neural networks to better understand their potential benefits. 

Taking into account the robustness to domain shift, it can be concluded that distillation is a better approach to train a network of a specific size than using pruning to obtain a smaller network from a larger one. Furthermore, the post-training quantization method is highly recommended over network pruning and KD techniques if compressing three-quarters of the model size is adequate for deployment, and no further training is required. This is due to the superior robustness of quantized models in dealing with domain shifts.

\subsection{Adversarial Attacks}

\begin{table*}[htb!]
    \centering
    \caption{Accuracies of both baseline-A and compressed ResNets under \textit{light} adversarial perturbations.}
    \resizebox{\textwidth}{!}{%
    \begin{tabular}{c|c|c|c|ccccccc|c}
    \toprule
    \begin{tabular}[c]{@{}c@{}} Base/Teacher \\ Model \end{tabular} &
    \begin{tabular}[c]{@{}c@{}} Compression \\ Method \end{tabular} & 
    \begin{tabular}[c]{@{}c@{}} \# Params \\ (M) \end{tabular} & 
    \begin{tabular}[c]{@{}c@{}} In-Domain \\ $\mathcal{D}_A$ Acc ($\%$) \end{tabular} & 
    \begin{tabular}[c]{@{}c@{}} DeepFool$_{L_\infty}$ \\ ($\epsilon=0.0003$) \end{tabular} & 
    \begin{tabular}[c]{@{}c@{}} PGD$_{L_\infty}$ \\ ($\epsilon=0.0003$) \end{tabular} & 
    \begin{tabular}[c]{@{}c@{}} FGSM \\ ($\epsilon=0.0003$) \end{tabular} & 
    \begin{tabular}[c]{@{}c@{}} C\&W$_{L_2}$ \\ ($\epsilon=0.03$) \end{tabular} & 
    \begin{tabular}[c]{@{}c@{}} DDN \\ ($\epsilon=0.003$) \end{tabular} & 
    \begin{tabular}[c]{@{}c@{}} EAD \\ ($\epsilon=1$) \end{tabular} &
    \begin{tabular}[c]{@{}c@{}} Salt\&Pepper \\ ($\epsilon=1$) \end{tabular} &
    \begin{tabular}[c]{@{}c@{}} Avg Adversarial \\ Acc ($\%$) \end{tabular} \\
    \midrule
    \multirow{2}{*}{ResNet-18} & - & 11.19 & 89.32 & 77.93 & 78.29 & 79.00 & 84.34 & 88.61 &  83.99 & 
     85.77 & 82.56 \\
     & FP & 4.51 &  89.68 & 82.21 & 82.56 & 83.63 & 86.48 & 87.54 & 85.77 & 85.77 & 84.85 \\
    \midrule
    \multirow{4}{*}{ResNet-34} & - & 21.30 & 90.75 & 76.87 & 76.51 & 78.29 & 83.63 & 89.32 & 81.49 & 84.70 & 81.54 \\
     & FP & 11.19 & 85.41 & 82.21 & 82.21 & 82.21 & 83.95 & 85.41 & 82.92 & 81.85 & 82.96 \\
     & FP & 4.51 & 75.44 & 71.17 & 71.17 & 71.17 & 73.31 & 75.44 & 71.17 & 70.82 & 72.03 \\
     & KD & 11.19 & 90.04 & 87.90 & 87.90 & 88.26 & 89.32 & 90.39 & 87.90 & 86.83 & \textbf{88.36} \\
    \midrule
    \multirow{6}{*}{ResNet-50} & - & 23.57 & 90.04 & 81.85 & 80.78 & 83.27 & 86.83 & 90.03 & 84.34 & 87.19 & 84.90 \\
     & FP & 21.30 & 92.17 & 85.05 & 84.34 & 85.41 & 87.19 & 91.46 & 86.12 & 89.32 & 86.98 \\
     & FP & 11.19 & 87.90 & 66.90 & 64.06 & 72.60 & 75.80 & 86.83 & 75.80 & 82.92 & 74.99 \\
     & FP & 4.51 & 75.44 & 49.11 & 45.20 & 54.80 & 63.34 & 74.02 & 59.07 & 68.68 & 59.17 \\
     & KD & 21.30 & 90.04 & 87.19 & 87.19 & 87.19 & 88.97 & 90.03 & 87.54 & 88.61 & \textbf{88.10} \\
     & KD & 11.19 & 90.39 & 87.54 & 87.54 & 87.54 & 89.32 & 90.75 & 88.26 & 87.90 & \textbf{88.41} \\
    \midrule
    \multirow{4}{*}{ResNet-101} & - & 42.56 & 91.81 & 80.07 & 79.00 & 80.78 & 85.05 & 91.10 & 83.99 & 87.90 & 83.98 \\
     & FP & 23.57 & 90.04 & 86.83 & 86.48 & 87.19 & 87.90 & 90.03 & 86.48 & 87.90 & 87.54 \\
     & FP & 11.19 & 86.12 & 81.49 & 81.14 & 83.27 & 84.70 & 86.12 & 82.92 & 83.27 & 83.27 \\
     & FP & 4.51 & 72.95 & 60.14 & 60.14 & 60.85 & 65.12 & 71.53 & 60.14 & 63.70 & 63.09 \\
    \midrule
    \multirow{5}{*}{ResNet-152} & - & 58.21 & 90.75 & 82.92 & 81.85 & 84.70 & 88.97 & 91.10 & 87.54 & 89.68 & 86.68 \\
     & FP & 42.56 & 91.46 & 81.85 & 80.43 & 83.98 & 85.76 & 91.46 & 85.41 & 88.61 & 85.36 \\
     & FP & 23.57 & 89.68 & 77.93 & 75.09 & 79.71 & 82.56 & 89.32 & 82.92 & 86.12 & 81.95 \\
     & FP & 11.19 & 82.56 & 60.14 & 57.29 & 63.70 & 68.32 & 81.49 & 67.97 & 72.95 & 67.41 \\
     & KD & 11.19 & 90.04 & 85.76 & 85.76 & 85.76 & 87.90 & 89.68 & 87.19 & 85.77 & \textbf{86.83} \\
     \bottomrule
    \end{tabular}%
    }
    \label{tab:adversarial_light}
\end{table*}

\begin{table*}[htb!]
\caption{Accuracies of both baseline-A and compressed ResNets under \textit{heavy} adversarial perturbations.}
    \centering
    \resizebox{\textwidth}{!}{%
    \begin{tabular}{c|c|c|c|ccccccc|c}
    \toprule
    \begin{tabular}[c]{@{}c@{}} Base/Teacher \\ Model \end{tabular} &
    \begin{tabular}[c]{@{}c@{}} Compression \\ Method \end{tabular} & 
    \begin{tabular}[c]{@{}c@{}} \# Params \\ (M) \end{tabular} & 
    \begin{tabular}[c]{@{}c@{}} In-Domain \\ $\mathcal{D}_A$ Acc ($\%$) \end{tabular} & 
    \begin{tabular}[c]{@{}c@{}} DeepFool$_{L_\infty}$ \\ ($\epsilon=0.004$) \end{tabular} & 
    \begin{tabular}[c]{@{}c@{}} PGD$_{L_\infty}$ \\ ($\epsilon=0.004$) \end{tabular} & 
    \begin{tabular}[c]{@{}c@{}} FGSM \\ ($\epsilon=0.004$) \end{tabular} & 
    \begin{tabular}[c]{@{}c@{}} C\&W$_{L_2}$ \\ ($\epsilon=0.4$) \end{tabular} & 
    \begin{tabular}[c]{@{}c@{}} DDN \\ ($\epsilon=0.24$) \end{tabular} & 
    \begin{tabular}[c]{@{}c@{}} EAD \\ ($\epsilon=10$) \end{tabular} 
    &
    \begin{tabular}[c]{@{}c@{}} Salt\&Pepper \\ ($\epsilon=15$) \end{tabular} 
    &
    \begin{tabular}[c]{@{}c@{}} Avg Adversarial \\ Acc ($\%$) \end{tabular} \\
    \midrule
    \multirow{2}{*}{ResNet-18} & - & 11.19 & 89.32 & 2.13 & 1.78 & 34.52 & 14.23 & 20.64 & 50.89 & 31.67 & 22.27 \\
     & FP & 4.51 & 89.68 & 20.28 & 11.73 & 49.47 & 38.43 & 48.75 & 62.28 &  16.73 & 35.38 \\
    \midrule
    \multirow{4}{*}{ResNet-34} & - & 21.30 & 90.75 & 3.91 & 0.36 & 47.69 & 14.95 & 18.51 & 49.47 & 50.89 & 26.54 \\
     & FP & 11.19 & 85.41 & 40.57 & 32.38 & 58.01 & 48.75 & 58.36 & 64.41 & 40.93 & 49.06 \\
     & FP & 4.51 & 75.44 & 30.60 & 21.71 & 46.95 & 40.21 & 49.82 & 48.04 & 21.71 & 37.01 \\
     & KD & 11.19 & 90.04 & 66.90 & 61.92 & 75.44 & 70.82 & 76.16 & 75.80 & 61.57 & \textbf{69.80} \\
    \midrule
    \multirow{6}{*}{ResNet-50} & - & 23.57 & 90.04 & 0.36 & 0.00 & 45.51 & 13.17 & 6.05 & 48.04 & 42.35 & 22.21 \\
     & FP & 21.30 & 92.17 & 6.76 & 1.78 & 53.74 & 22.06 & 26.33 & 57.30 & 55.16 & 31.88 \\
     & FP & 11.19 & 87.90 & 0.71 & 0.35 & 32.38 & 4.63 & 1.78 & 37.37 & 40.57 & 16.83 \\
     & FP & 4.51 & 75.44 & 0.00 & 0.00 & 9.25 & 0.71 & 0.00 & 17.08 & 12.46 & 5.64 \\
     & KD & 21.30 & 90.04 & 69.03 & 64.06 & 73.31 & 70.82 & 77.58 & 76.87 & 66.55 & \textbf{71.17} \\
     & KD & 11.19 & 90.39 & 62.63 & 58.72 & 71.89 & 67.97 & 75.80 & 75.44 & 49.11 & \textbf{65.94} \\
    \midrule
    \multirow{4}{*}{ResNet-101} & - & 42.56 & 91.81 & 6.05 & 1.78 & 59.07 & 12.10 & 11.39 & 57.30 & 64.41 & 30.30 \\
     & FP & 23.57 & 90.04 & 41.28 & 27.76 & 66.19 & 53.38 & 62.28 & 70.46 & 56.23 & 53.94 \\
     & FP & 11.19 & 86.12 & 28.47 & 17.44 & 49.82 & 43.42 & 53.38 & 56.94 & 27.40 & 39.55 \\
     & FP & 4.51 & 72.95 & 2.13 & 0.71 & 13.52 & 10.68 & 12.81 & 19.93 & 6.05 & 9.40 \\
    \midrule
    \multirow{5}{*}{ResNet-152} & - & 58.21 & 90.75 & 7.12 & 4.98 & 63.34 & 15.30 & 12.46 & 61.92 & 65.12 & 32.89 \\
     & FP & 42.56 & 91.46 & 5.69 & 1.07 & 61.56 & 9.96 & 7.12 & 56.23 & 52.67 & 27.76 \\
     & FP & 23.57 & 89.68 & 1.42 & 0.71 & 55.87 & 7.83 & 5.34 & 50.53 & 35.59 & 22.47 \\
     & FP & 11.19 & 82.56 & 0.35 & 0.00 & 29.54 & 4.27 & 1.07 & 22.42 & 7.12 & 9.25 \\
     & KD & 11.19 & 90.04 & 69.75 & 65.48 & 75.80 & 72.95 & 78.65 & 77.94 & 53.74 & \textbf{70.62} \\
     \bottomrule
    \end{tabular}%
    }
    \label{tab:adversarial_heavy}
\end{table*}

The accuracy of the quantized ResNets and their corresponding baselines against EAD, C\&W, and Salt\&Pepper adversarial attacks are demonstrated in \cref{tab:adversarial_qt}. Our results indicate that while the gradient-based attacks, EAD and C\&W, significantly degrade the robustness of baselines, these attacks barely reduce the accuracy of the quantized models. For example, the accuracy of the uncompressed ResNet-18 drops significantly from 88.61\% on the clean data to 50.89\% and 0.00\% under heavy EAD and C\&W attacks, respectively. However, the accuracy of the quantized ResNet-18 is reduced from 88.26\% on the clean data to 87.90\% under the gradient-based attacks. This observation shows the comparative robustness of quantized models against gradient-based attacks due to the phenomenon of \textit{gradient masking} \cite{galloway2018attacking,papernot2017practical,li2017pruning}, which hinders the attackers' ability to find gradients that can mislead the quantized model. However, adversarial attacks can be specifically designed to overcome gradient masking \cite{athalye2018obfuscated} and challenge the quantized models.

While our experiments in \cref{tab:adversarial_qt} revealed that quantized models are vulnerable to non-gradient-based attacks such as Salt\&Pepper, we also observed that quantization improves the accuracy of smaller models (\ie, ResNet-18 and ResNet-50) compared to their baselines under heavy Salt\&Pepper attacks. However, deeper models such as ResNet-152 experienced a reduction in accuracy after quantization.%, suggesting that the benefits of quantization may depend on the model's architecture and complexity.

%Nevertheless, according to our results provided in \cref{tab:adversarial_qt}, the quantized models are vulnerable to non-gradient-based attacks like Salt\&Pepper. Under heavy Salt\&Pepper attacks, quantization improves the accuracy of smaller models (\ie, ResNet-18 and ResNet-50) when compared to the baselines, deeper models (\ie, ResNet-152) experience a reduction in accuracy after quantization. This can be explained by the \textit{error amplification effect} \cite{liao2018defense,lin2018defensive}, which occurs when the perturbation is magnified by the quantization operation through the layers of a DNN. 

% the average accuracy of each quantized model is about $\mathbf{50\%}$ \textbf{higher} than of the corresponding baseline. Therefore, the quantized models are highly robust to adversarial attacks, outperforming the baselines by a significant gap. Furthermore, the performance of the quantized models under heavy and light attacks is roughly the same, while the accuracy of baselines degrades considerably. This observation is consistent with the findings of \cite{krizhevsky2012imagenet}. One possible explanation for this is that the quantization of activations and weights reduces the model's sensitivity to small perturbations imposed by adversarial attacks, making them less effective on quantized models. 

We also investigate the robustness of models that are compressed using pruning and KD against light and heavy adversarial attacks, and the results are presented in \cref{tab:adversarial_light,tab:adversarial_heavy}. In contrast to the quantized models, the pruned and distilled models are vulnerable to all types of attacks. Comparing the accuracy of the pruned models with their corresponding baselines, we observe that, with the exception of ResNet-152, the pruned models with the lowest pruning rate outperform the baselines under both light and heavy attacks. However, the larger size of ResNet-152 and, consequently, the larger amount of pruned parameters, may explain why pruning does not improve the adversarial robustness of this model. We also notice that the average adversarial robustness decreases by increasing the pruning rate. 
% Therefore, low pruning rates improve the adversarial robustness of relatively smaller models, while larger models do not benefit from pruning.

Furthermore, as shown in both \cref{tab:adversarial_light,tab:adversarial_heavy}, the compressed models that use KD consistently outperform both their baselines/teacher models and pruned models of the same size with a significant performance gap between 20\% and 60\% under heavy attacks. Moreover, the adversarial robustness of a ResNet-18, trained by distillation from teachers with different sizes (\ie, ResNet-34, ResNet-50, and ResNet-152) remains roughly the same against both light and heavy attacks. This suggests that, unlike pruning, KD can maintain the model's adversarial robustness across different compression rates.  

A comparison between the quantized models and pruned/distilled ones is not quite fair since they have different architectures. However, according to our results in \cref{sec:domain_shifts}, quantized models performed the best on domain shift experiments. Also, according to the accuracy of compressed models under Salt\&Pepper, quantized models achieved the best results.

% Although the quantized models are evaluated against only a subset of studied attacks (see \cref{sec:adv}), by comparing the accuracy of quantized models with the pruned and distilled ones under C\&W and EAD attacks, we conclude that quantized models are significantly more robust against adversarial attacks. Moreover, KD maintains a convenient balance between adversarial robustness and computational complexity by generating smaller, yet robust models.
% Note that the adversarial results are in-line with domain shift results. 

%----------------------------------------------------------
\section{Conclusion}
This paper explores the robustness of compressed networks to various distribution shifts using the Office-31 dataset. We observe that the compressed models perform worse in the unseen domain as the compression ratio increases due to distribution shifts. The results indicate that as the compression ratio increases, the compressed models perform more poorly in the unseen domain due to distribution shifts. Furthermore, we discover that compressed networks originating from smaller models demonstrate better generalization abilities in the target domain, indicating that they are more robust to distribution shifts compared to networks that were originally as large.
 
In terms of compression techniques for neural networks, the pruning technique is known for generating highly sensitive compressed networks that are vulnerable to domain shifts and adversarial perturbations. On the other hand, compact networks produced through knowledge distillation are less affected by these issues. It is worth emphasizing that quantized networks, which are compressed to approximately 25\% of their original size, offer significantly more robustness to distribution shifts, particularly in the case of domain shifts, than other compressed networks.

%----------------------------------------------------------
%%%%%%%%% REFERENCES
{\small
\bibliographystyle{IEEEtran}
\bibliography{IEEEabrv,main.bib}
}

\newpage
\onecolumn
\end{document}